\documentclass[final]{cvpr}

\usepackage{times}
\usepackage{epsfig}
\usepackage{graphicx}
\usepackage{amsmath}
\usepackage{amssymb}
\usepackage{multirow}
\usepackage{nopageno}

\usepackage{enumitem} 

\DeclareMathOperator*{\argmin}{argmin}
\DeclareMathOperator*{\argmax}{argmax}

\usepackage{booktabs}
\usepackage{subfigure}
\usepackage{caption}
\usepackage{pifont}
\usepackage{soul}
\usepackage[pagebackref=true,breaklinks=true,colorlinks,bookmarks=false]{hyperref}

\def\cvprPaperID{5886} 
\def\confYear{CVPR 2021}
\renewenvironment{quote}{
	\onecolumn
}

\begin{document}
	\begin{quote}
		
		\newpage
		\onecolumn
		\noindent \vspace{1cm}
		\noindent \textbf{\huge{Learning Normal Dynamics in Videos with Meta Prototype Network}}
		
		\vspace{2cm}
		
		\noindent {\LARGE{Hui Lv, Chen Chen, Zhen Cui, Chunyan Xu, Yong Li, Jian Yang}}
		
		\vspace{2cm}
		
		
		\vspace{1cm}
		
		\noindent For reference of this work, please cite:
		
		\vspace{1cm}
		\noindent Hui Lv, Chen Chen, Zhen Cui, Chunyan Xu, Yong Li, Jian Yang.
		Learning Normal Dynamics in Videos with Meta Prototype Network. \emph{Proceedings of the IEEE International Conference on Computer Vision and Pattern Recognition.} 2021.
		
		\vspace{1cm}
		
		\noindent Bib:\\
		\noindent
		@inproceedings\{Lv2021MPN,\\
		\ \ \   title=\{Learning Normal Dynamics in Videos with Meta Prototype Network\},\\
		\ \ \  author=\{Lv, Hui and Chen, Chen and Zhen, Cui and Xu, Chunyan and Yang, Jian\},\\
		\ \ \  booktitle=\{Proceedings of the IEEE International Conference on Computer Vision and Pattern Recognition\},\\
		\ \ \  year=\{2021\}\\
		\}
	\end{quote}

	\title{Learning Normal Dynamics in Videos with Meta Prototype Network}
	
	\twocolumn
	\author{Hui Lv$^1$, Chen Chen$^2$, Zhen Cui$^1$\thanks{Corresponding authors}, Chunyan Xu$^1$, Yong Li$^1$, Jian Yang$^1$\\
		\small{$^1$PCALab, Nanjing University of Science and Technology,
			$^2$University of North Carolina at Charlotte}\\
		{\tt\small \{hubrthui, zhen.cui, cyx, yong.li, csjyang\}@njust.edu.cn, chen.chen@uncc.edu}
	}
	
	\maketitle

	\begin{abstract}
		Frame reconstruction (current or future frame) based on Auto-Encoder (AE) is a popular method for video anomaly detection. 
		With models trained on the normal data, the reconstruction errors of anomalous scenes are usually much larger than those of normal ones. 
		Previous methods introduced the memory bank into AE, for encoding diverse normal patterns across the training videos.
		However, they are memory-consuming and cannot cope with unseen new scenarios in the testing data.
		In this work, we propose a dynamic prototype unit (DPU) to encode the normal dynamics as prototypes in real time, free from extra memory cost. 
		In addition, we introduce meta-learning to our DPU to form a novel few-shot normalcy learner, namely Meta-Prototype Unit (MPU). It enables the fast adaption capability on new scenes by only consuming a few iterations of update. 
		Extensive experiments are conducted on various benchmarks. The superior performance over the state-of-the-art demonstrates the effectiveness of our method. \textbf{Our code is available at} \url{https://github.com/ktr-hubrt/MPN/}.
	\end{abstract}
	
	\section{Introduction}
	Video anomaly detection (VAD) refers to the identification of behaviors or appearance patterns that do not conform to the expectation \cite{adam2008robust,benezeth2009abnormal,chandola2009anomaly,lv2020localizing}. 
	Recently, there is a growing interest in this research topic because its key role in surveillance for public safety, \eg the task of monitoring video in airports, at border crossings, or at government facilities becomes increasingly critical. 
	However, the `anomaly' is conceptually unbounded and often ambiguous, making it infeasible to gather data of all kinds of possible anomalies. 
	Anomaly detection is thus typically formulated as an unsupervised learning problem, aiming at learning a model to exploit the regular patterns only with the normal data. During inference, patterns that do not agree with the encoded regular ones are considered as anomalies.
	\begin{figure}[t]
		\centering
		\includegraphics[width= 0.45\textwidth]{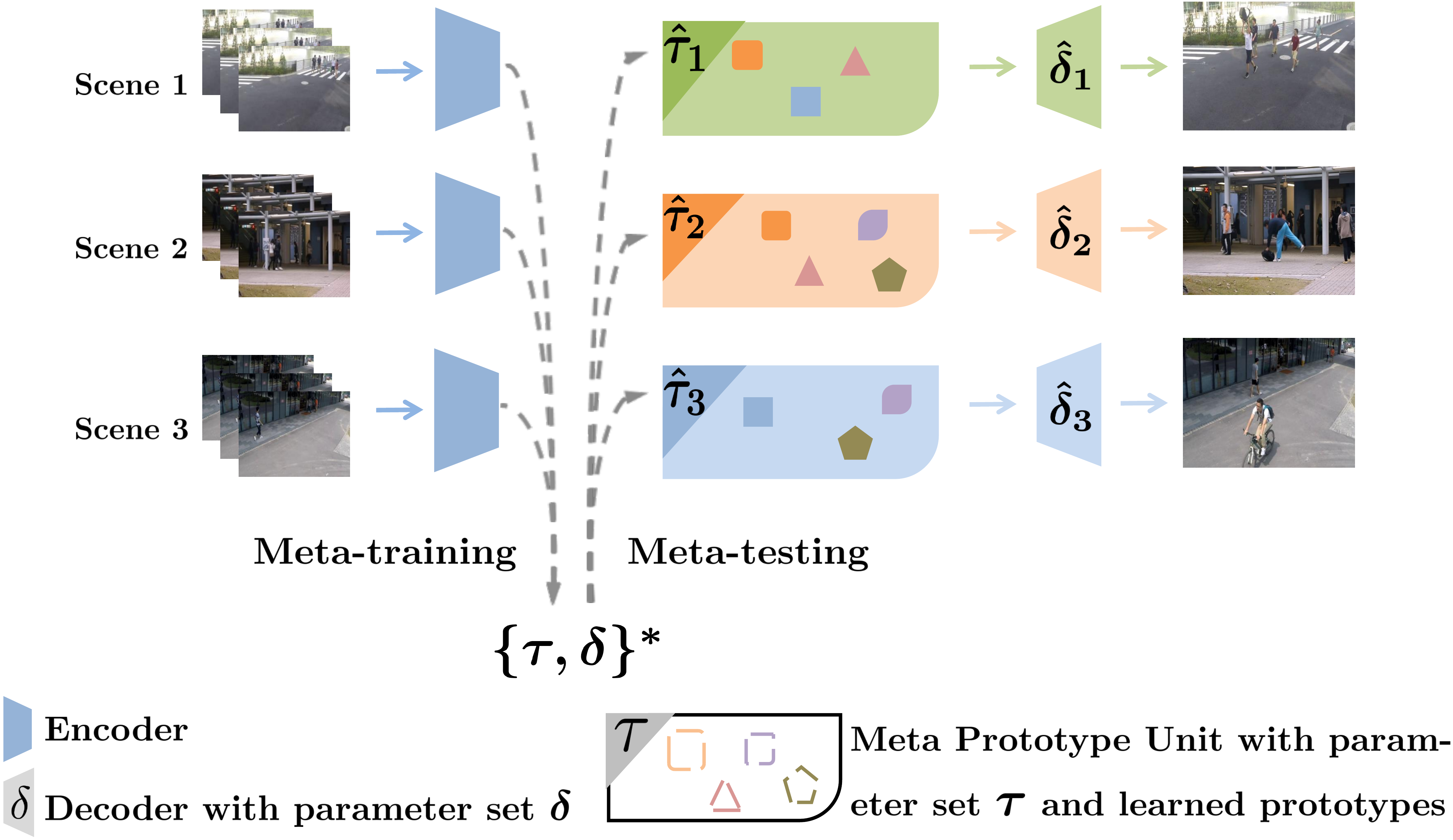}
		\caption{An overview of our approach. 
			(1) We design a Dynamic Prototype Unit (DPU) to learn a pool of prototypes for encoding normal dynamics;
			(2) Meta-learning methodology is introduced to formulate the DPU as a few-shot normalcy learner -- Meta Prototype Unit (MPU). It improves the scene adaption capacity by learning an initialization of the target model and adjusting it to new scenes with parameters update during inference. Better viewed in color.}
		\label{fig:MPU}
	\end{figure}
	
	Deep Auto-Encoder (AE)~\cite{ronneberger2015u} is a popular approach for video anomaly detection. 
	Researchers usually adopt AEs to model the normal patterns with historical frames and to reconstruct the current frame \cite{hasan2016learning,masci2011stacked,sabokrou2016video,chalapathy2017robust,sabokrou2018adversarially,abati2019latent} or predict the upcoming frame \cite{liu2018future,nguyen2019anomaly,lu2019future,luo2017remembering,gong2019memorizing}.
	For simplicity, we refer to the two cases as frame prediction. 
	Since the models are trained with only normal data, higher prediction errors are expected for abnormal (unseen patterns) inputs than those of the normal counterparts.
	Previously, many methods are based on this assumption for anomaly detection.
	However, this assumption does not always hold true.
	
	On the one hand, the existing methods rely on large volumes of normal training data to model the shared normal patterns.
	These models are prone to face the `over-generalizing' dilemma, where all video frames can be predicted well, no matter they are normal or abnormal, owing to the powerful representation capacity of convolutional neural networks (CNNs) \cite{park2020learning,gong2019memorizing}.
	Previous approaches \cite{park2020learning,gong2019memorizing} proposed to explicitly model the shared normal patterns across normal training videos with a memory bank, for the propose of boosting the prediction of normal regions in frames while suppressing the abnormal ones. 
	However, it is extremely memory-consuming for storing the normal patterns as memory items across the whole training set.
	
	To tackle this limitation, we propose to encode the normal dynamics in an attention manner, which is proven to be effective in representation learning and enhancement \cite{vaswani2017attention, li2019selective, hu2018squeeze}.
	A normalcy learner, named as Dynamic Prototype Unit (DPU), is developed to be easily incorporated into the AE backbone. It
	takes the encoding of consecutive normal frames as input, then learns to mine diverse normal dynamics as compact prototypes. 
	More specifically, we apply a novel attention operation on the AE encoding map, which assigns a normalcy weight to each pixel location to form a normalcy map.
	Then, prototypes are obtained as an ensemble of the local encoding vectors under the guidance of normalcy weights.
	Multiple parallel attention operations are applied to generate a pool of prototypes.
	With the proposed compactness and diverseness feature reconstruction loss function, the prototype items are trained to represent diverse and compact dynamics of the shared normal patterns in an end-to-end fashion.
	Finally, the AE encoding map is aggregated with the normalcy encoding reconstructed by prototypes for latter frame prediction.
	
	On the other hand, the normal patterns appearing in various scenes differ from each other. For instance, a person running in a walking zone is regarded as an anomaly, while this activity is normal in the playground. 
	Previous methods~\cite{liu2018future,gong2019memorizing} assume the normal patterns in training videos are consistent with those of test scenes in the unsupervised setting of VAD. 
	However, this assumption is unreliable, especially in real-world applications where surveillance cameras are installed in various places with significantly different scenarios. Therefore, there is a pressing need to develop an anomaly detector with adaption capability.
	To this end, \cite{park2020learning} defines a rule for updating items in the memory bank based on a threshold to record normal patterns and ignore abnormal ones.
	However, it is impossible to find a uniform and optimal threshold for distinguishing the normal and abnormal frames under various scenarios.

	In this work, we approach this problem from a new perspective, 
	motivated by~\cite{lu2020few}, which is the few-shot setting for video anomaly detection.
	In the few-shot setting, videos from multiple scenes are accessible during training, and a few video frames from target scene are available during inference.
	A solution to this problem is using the meta-learning technique. 
	In this meta-training phase, a few-shot target model is trained to adapt to a new scene with a few frames and parameters update iterations.
	The procedure is repeated using video data from different scenes for obtaining a model initialization that serves as a good starting point for fast adaption to new scenes.
	Therefore, we formulate our DPU module as a few-shot normalcy learner, namely Meta Prototype Unit (MPU), for the goal of learning to learn the normalcy in target scenes.
	Rather than roughly shifting to the new scene by adjusting the whole network~\cite{lu2020few}, which may lead to the `over-generalizing' problem, we propose to freeze the pre-trained AE and only update the parameters of our MPU.
	Consuming only a few parameters and update iterations, our meta-learning model is endued with the power of fast and effective adaption to the normalcy of unseen scenarios. An overview of our approach is presented in Fig. \ref{fig:MPU}. 
		
	We summarize our contributions as follows:
	i) We develop a Dynamic Prototype Unit (DPU) for learning to represent diverse and dynamic patterns of the normal data as prototypes. An attention operation is thus designed for aggregating the normal dynamics to form prototype items.
	The whole process is differentiable and trained end-to-end.
	ii) We introduce meta-learning into our DPU and improve it as a few-shot normalcy learner -- Meta Prototype Unit (MPU). It effectively endows the model with the fast adaption capability by consuming only a few parameters and update iterations.
	iii) Our DPU-based AE achieves new state-of-the-art (SOTA) performance on various unsupervised anomaly detection benchmarks. In addition, experimental results validate the adaption capability of our MPU in the few-shot setting.
	
	\section{Related Work}
	\noindent \textbf{Anomaly Detection.}
	Due to the absence of anomaly data and expensive costs of annotations, video anomaly detection has been formulated into several types of learning problems.
	For example, the unsupervised setting assumes only normal training data~\cite{li2013anomaly,luo2017revisit,lu2013abnormal}, and weakly-supervised setting can access videos with video-level labels~\cite{sultani2018real,zhong2019graph,lv2020localizing}. 
	In this work, we focus on the unsupervised setting, which is more practical in real applications. For example, the normal video data of surveillance cameras are easily accessible for learning models describing the normality.
	Earlier methods, based on sparse coding~\cite{cong2011sparse,zhao2011online,lu2013abnormal}, markov random field~\cite{kim2009observe}, a mixture of dynamic textures~\cite{mahadevan2010anomaly}, a mixture of probabilistic
	PCA models~\cite{5206569}, \etc., tackle the task as a novelty detection problem~\cite{lv2020localizing}.
	Latter, deep learning (CNNs in particular) has triumphed over many computer vision tasks including video anomaly detection (VAD).
	In~\cite{luo2017revisit}, Luo~\emph{et al}. propose a temporally coherent sparse coding-based method which can be mapped to a stacked RNN framework. 
	
	Recently, many methods leverage deep Auto-Encoder (AE) to model regular patterns and reconstruct video frames~\cite{hasan2016learning,masci2011stacked,sabokrou2016video,chalapathy2017robust,sabokrou2018adversarially,abati2019latent}. 
	Multiple variants of AE have been developed to cooperate spatial and temporal information for video anomaly detection. In \cite{luo2017remembering,chong2017abnormal}, the authors investigate Recurrent Neural Network (RNN) and Long Short Term Memory (LSTM) for modeling regular patterns in sequential data. 
	Liu~\emph{et al}.~\cite{liu2018future} propose to predict the future frame with AE and Generative Adversarial Network (GAN).
	They assume anomalous frames are unpredictable in the video sequence.
	It has achieved superior performance over previous reconstruction-based methods.
	However, this kind of methodology suffers from the `over-generalizing' problem that sometimes anomalous frames can also be predicted well (\ie small prediction error) as normal ones.
	
	Gong~\emph{et al}. (MemAE)~\cite{gong2019memorizing} and Park~\emph{et al}. (LMN)~\cite{park2020learning} introduce a memory bank into the AE for anomaly detection. 
	They record normal patterns across training videos as memory items in a bank, which brings extra memory cost. 
	While we propose to learn the normalcy with an attention mechanism to measure the normal extent.
	The learning procedure is fully differentiable and the prototypes are dynamically learned with the benefits of adapting to the current scene
	spatially and temporally, compared with querying and updating the memory bank with pre-defined rules for recording rough patterns cross the training data in~\cite{gong2019memorizing, park2020learning}.
	Moreover, the prototypes are automatically derived based on the real-time video data during inference, without referencing to the memory items collected from the training phase~\cite{gong2019memorizing,park2020learning}.
	For adaption to test scenes, Park~\emph{et al}.~\cite{park2020learning} further expand the update rules of the memory bank by using a threshold to distinguish abnormal frames and record normal patterns.
	However, it is impossible to find a uniform and optimal threshold for distinguishing the normal and abnormal frames under various scenarios.
	On the contrary, we introduce the meta-learning technology into our DPU module to enable the fast adaption capacity to a new scenery.
	
	\noindent \textbf{Attention Mechanisms.}
	Attention mechanism~\cite{wang2017residual,woo2018cbam,hu2018squeeze,su2019multi,kligvasser2018xunit,yu2018learning,fu2019dual,zhao2018psanet,li2019selective} is widely adopted in many computer vision tasks. 
	Current methods can be roughly divided into two categories, which are the channel-wise attention~\cite{woo2018cbam,hu2018squeeze,su2019multi,yu2018learning} and spatial-wise attention~\cite{su2019multi,zhao2018psanet,woo2018cbam,kligvasser2018xunit,fu2019dual}.
	SENet~\cite{hu2018squeeze} designs an effective and lightweight gating mechanism to self-recalibrate the feature map via channel-wise importance. 
	Wang~\emph{et al}.~\cite{wang2017residual} propose a trunk-and-mask attention between intermediate stages of a CNN.
	However, most prior attention modules focus on optimizing the backbone for feature learning and enhancement. We propose to leverage the attention mechanism to measure the normalcy of spatial local encoding vectors, and use them to generate prototype items which encode the normal patterns.
	
	\noindent \textbf{Few-Shot and Meta-learning.}
	In few-shot learning, researchers aim to mimic the fast and nimble learning ability of humans, which can quickly adapt to a new scenario with only a few data examples~\cite{lake2015human}.
	Generally, meta-learning has been developed to tackle this problem.
	The meta-learning methods mainly fall into three categories: metric-based~\cite{koch2015siamese,vinyals2016matching,sung2018learning}, model-based~\cite{santoro2016meta,munkhdalai2017meta} and optimization-based approaches~\cite{finn2017model}. 
	These methods can quickly adapt to a new task through the meta-update scheme among multiple tasks during parameter optimization. However, most of the approaches above are designed for simple tasks like image classification. Recently, Lu~\emph{et al}.~\cite{lu2020few} follow the optimization-based meta-learning approach \cite{finn2017model} and apply it to train a model for scene-adaptive anomaly detection. They simply set the whole network as the few-shot target model for meta-learning, for learning an initialization parameter set of the entire model.
	However, in this work, we learn two sets of initial parameters and update step sizes separately for an elaborate updating of designed module in our model with fewer parameters and update iterations.
	
	\section{Method}
	In this section, we elaborate the proposed method for VAD.
	First, we describe the learning process of normal dynamics in the Dynamic Prototype Unit (DPU) in Sec.~\ref{sec_dpu},
	and we explain the objective functions of the framework in Sec.~\ref{objf}.
	Then in Sec.~\ref{MFVAD}, we present the details of the few-shot normalcy learner. 
	Finally in Sec.~\ref{vad_pipe}, we detail the training and testing procedures of our VAD framework.
	\begin{figure}[t]
		\centering
		\includegraphics[width= 0.4\textwidth]{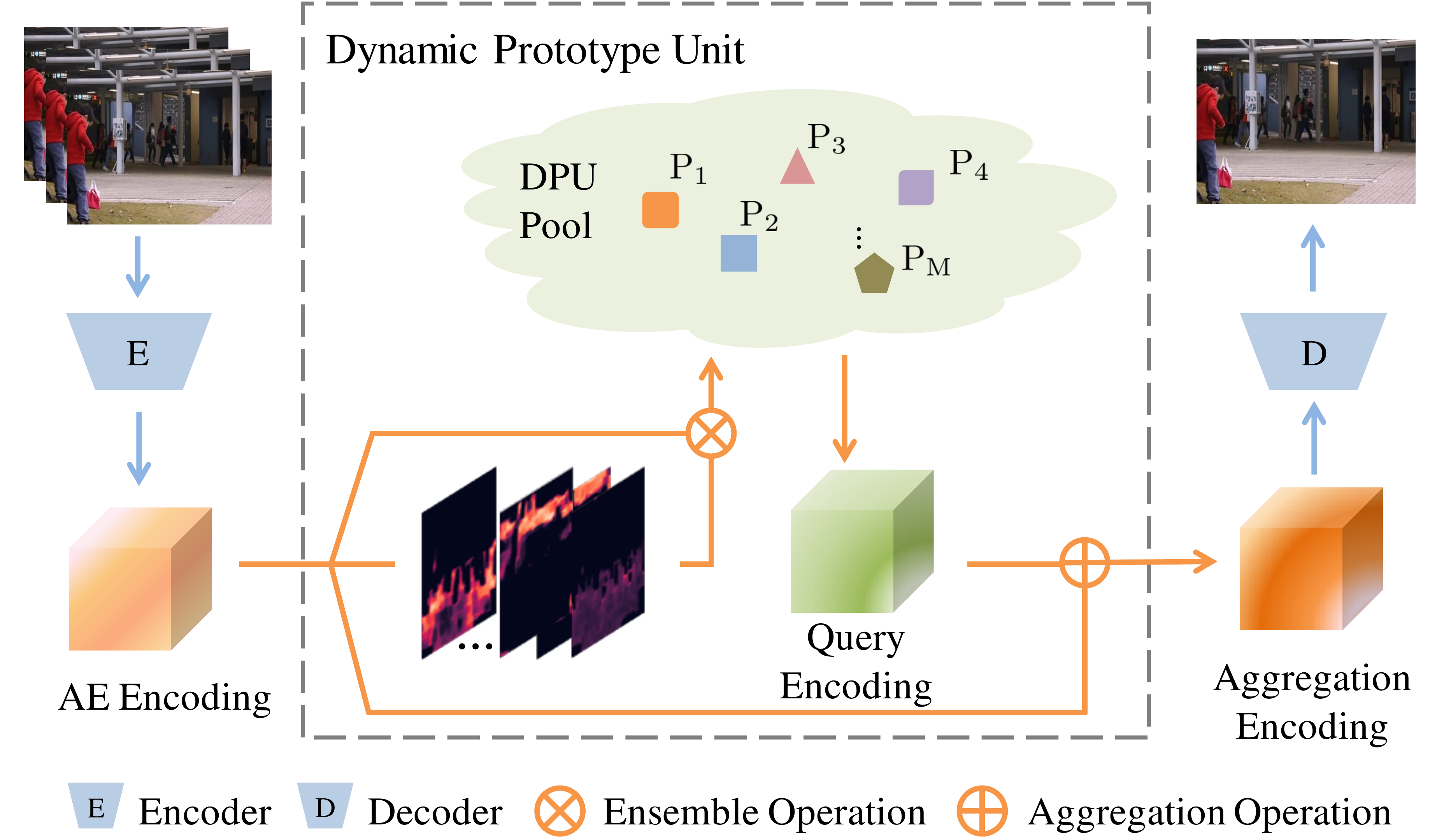}
		\caption{The framework of DPU-based model. The proposed Dynamic Prototype Unit (DPU) is plugged into an Auto-Encoder (AE) to learn prototypes for encoding normal dynamics. The prototypes are obtained from the AE encoding with the guidance of normalcy weights and the normalcy weights of the AE encoding are generated in a fully differentiable attention manner. Then an normalcy encoding map (green color) is reconstructed as an encoding of learned prototypes. It is further aggregated with the AE encoding map for latter frame prediction.}
		\label{fig:framework}
	\end{figure}
	
	\subsection{Dynamic Prototype Unit}
	\label{sec_dpu}
	The framework of the DPU-based AE is shown in Fig.~\ref{fig:framework}. 
	DPU is trained to learn and compress normal dynamics of real-time sequential information as multiple prototypes and enrich the input AE encoding with normal dynamics information.
	Note that, DPU can be plugged into different places (with different resolutions) of the AE.  We conduct ablation studies in Sec.~\ref{MPU_R} to analyze the impact of DPU position.
	
	Let's first consider an AE model takes as inputs the $\rm T$ observed video frames $(\rm I_{k-{\rm T}+1},I_{k-{\rm T}+2},...,I_k)$, simplified as $x_k$. 
	Then the selected hidden encoding of AE is feed forward into our DPU $P_{\tau}: \mathbb{R}^{\rm{h \times w \times c}} \rightarrow \mathbb{R}^{\rm{h \times w \times c}}$.
	Finally, the output encoding of DPU is run through the remaining AE layers (after DPU) for predicting upcoming ground-truth frame $y_k = {\rm I_{k+1}}$.
	We denote the frame sequence as an input$\&$output pair $(x_k,y_k)$ of the $k$-th moment.
	
	The forward pass of DPU is realized by generating a pool of dynamic prototypes in a fully differentiable attention manner, then reconstructing a normalcy encoding by retrieving the prototypes, and eventually aggregating the input encoding with the normalcy encoding as the output.
	The whole process can be broken down into 3 sub-processes, which are $Attention$, $Ensemble$ and $Retrieving$. 
	
	Concretely, the $\rm t$-th input encoding map ${\mathcal X}_{\rm t} = f_\theta(x_{\rm t}) \in \mathbb{R}^{\rm{h,w,c}}$ from AE is first extracted, viewed as $\rm N=w*h$ vectors of $c$ dimensional, $\{\rm x_t^1, x_t^2,...,x_t^N\}$. 
	In the sub-process of $Attention$, a quantity of $M$ attention mapping functions $\{\psi_m: \mathbb{R}^{\rm{c}} \rightarrow \mathbb{R}^{\rm{1}}\}_{m=1}^{M}$ are employed to assign normalcy weights to encoding vectors, $w_{\rm t}^{n,m} \in {\mathcal W}_{\rm t}^m = \psi_{m}({\mathcal X}_{\rm t})$.
	On each pixel location, the normalcy weight measures the normalcy extent of the encoding vector. 
	Here, ${\mathcal W}_{\rm t}^m \in \mathbb{R}^{\rm{h \times w \times 1}}$ denotes the $m$-th normalcy map, generated from the $m$-th attention function.
	Then one unique prototype ${\rm p}_{\rm t}^{m}$ is derived as an ensemble of $\rm{N}$ encoding vectors with normalized normalcy weights in sub-process $Ensemble$ as:
	\begin{equation}
	{\rm p}_{\rm t}^{m} = \sum\limits_{n=1}^{\rm{N}} \frac{w_{\rm t}^{n,m}}{\sum_{{n}'=1}^{\rm{N}} {w}_{\rm t}^{n',m}} {\rm x}_{\rm t}^{n}.
	\end{equation}
	Similarly, $M$ prototypes are derived from multiple attention functions to form a prototype pool, ${{\mathcal P}_{\rm t}} = \{{\rm p}_{\rm t}^{m}\}_{m=1}^{M}$.
	
	Finally, in the $Retrieving$ sub-process, input encoding vectors ${\rm x}_{\rm t}^n$ ($n \in {\rm N}$) from the AE encoding map are used as queries to retrieve relevant items in the prototype pool for reconstructing a normalcy encoding $\widetilde{\rm X}_{\rm t} \in \mathbb{R}^{\rm{h \times w \times c}}$. 
	For every obtained normalcy encoding vector, this proceeds as:
	\begin{equation}
	\tilde{\rm{x}}_{\rm t}^n = \sum\limits_{m=1}^M \beta_{\rm t}^{n,m}{\rm p}_{\rm t}^{m},
	\label{eq:2}
	\end{equation}
	where $\beta_{\rm t}^{n,m} = \frac{{\rm x}_{\rm t}^{n} {\rm p}_{\rm t}^m}{\sum_{m'=1}^M {\rm x}_{\rm t}^{n} {\rm p}_{\rm t}^{m'}}$ denotes the relevant score between the $n$-th encoding vector ${\rm x}_{\rm t}^n$ and the $m$-th prototype item ${\rm p}_{\rm t}^{m}$.
	The obtained normalcy map is aggregated with the original encoding ${\mathcal X}$ as the final output using a channel-wise sum operation.
	The key idea is to enrich the AE encoding with the normalcy information to boost the prediction of normal parts of video frames while suppressing the abnormal parts. The output encoding of DPU goes through the remaining AE layers for later frame prediction.

	\subsection{VAD Objective Functions}
	\label{objf}
	In this section, we present the objective functions in the pipeline, which enable the prototype learning for normalcy dynamics representation, feature reconstruction for normalcy enhanced encoding, and frame prediction for anomaly detection.
	To train our model, the overall loss function $\mathcal{L}$ consists of a feature reconstruction term $\mathcal{L}_\mathrm{fea}$ and a frame prediction term $\mathcal{L}_\mathrm{fra}$. These two terms are balanced by weight $\lambda_{1}$ as:
	\begin{equation}
	\label{Overall_L}
	\mathcal{L} = \mathcal{L}_\mathrm{fra} + \lambda_\mathrm{1}\mathcal{L}_\mathrm{fea}.
	\end{equation}
	
	\noindent \textbf{Frame Prediction Loss} is formulated as the L2 distance between ground-truth $y_{\rm t}$ and network prediction $\hat{y}_{\rm t}$:
	\begin{equation}
	\label{FPL}
	\mathcal{L}_\mathrm{fra} =  \|\hat{y}_{\rm t}-{y}_{\rm t}\|_2.
	\end{equation}
	
	\noindent \textbf{Feature Reconstruction Loss} is designed to make the learned normal prototypes have the properties of compactness and diversity. It has two terms $\mathcal{L}_\mathrm{c}$ and $\mathcal{L}_\mathrm{d}$, aiming at the two properties respectively, and is written as:
	\begin{equation}
	\label{fea_re}
	\mathcal{L}_\mathrm{fea} =  \mathcal{L}_\mathrm{c} + \lambda_\mathrm{2}\mathcal{L}_\mathrm{d},
	\end{equation}
	where $\lambda_\mathrm{2}$ is the weight parameter. The compactness term $\mathcal{L}_\mathrm{c}$ is for reconstruction of normalcy encoding with compact prototypes. It measures the mean L2 distance of input encoding vectors and their most-relevant prototypes as:
	\begin{align}
	\mathcal{L}_\mathrm{c} &=\frac{1}{\rm N}\sum_{n=1}^{\rm N} \|{\rm x}_{\rm t}^n-{\rm p}_{\rm t}^{*}\|_2,\\
	&{\rm s.t.,}~* = \argmax_{m \in [1, {M}]} \beta_{\rm t}^{n,m},
	\end{align}
	where $\beta^{n,m}$ is the relevant score mentioned in Eq.~\ref{eq:2}. Note that, $\argmax$ is only used to obtain indices of the most relevant vector, and not involved
	in the back-propagation.
	We further promote the diversity among prototype items by pushing the learned prototypes away from each other. The diversity term $\mathcal{L}_\mathrm{d}$ is expressed as:
	\begin{align}
	\mathcal{L}_\mathrm{d} =\frac{2}{{M(M-1)}} \sum_{m=1}^{M} \sum_{m'=1}^{M} \lbrack-||p_m-p_{m'}||_2+\gamma \rbrack_{+}.
	\end{align}
	Here, $\gamma$ controls the desired margin between prototypes. Taking benefits of above two terms, the prototype items are encouraged to encode compact and diverse normalcy dynamics for normal frame prediction.
	
	\subsection{Meta-learning in Few-shot VAD}
	\label{MFVAD}
	Generally, the AEs take consecutive video frames as inputs and reconstruct the current frame or predict the subsequent frame.
	In this work, we focus on the latter paradigm.
	We first consider a VAD architecture formulated as $f_{\theta}(E_{\eta}({x})) = D_{\delta}(P_{\tau}(E_{\eta}({x})))$, where $\eta$, $\delta$ denote the parameters of the AE encoding/decoding function $E$, $D$, respectively.
	The designed model takes as input a sequence of frame samples $x$. 
	Then the AE encoding ${\mathcal X} = E_{\eta}({x})$ is fed into the DPU module $P_{\tau}$.
	DPU learns to encode the normal dynamics information in consecutive video frames with the parameter set $\tau$.
	Our few-shot target model $f_{\theta}({\mathcal X})$, namely Meta-Prototype Unit (MPU), consists of the main module DPU and the AE decoder with parameter set $\theta = \tau \cup \delta$.
	Taking the subsequent frame sample $y$ as the ground-truth, the target model is updated based on the objective functions defined in Sec.~\ref{objf}.
	The process is denoted as the update function $U$ with frame pair $(x,y)$.
	
	During inference, short normal clips of test videos are available for adjusting the model to the new scenery in the few-shot setting of VAD. 
	To mimic this adaption process, meta-training strategy is implemented in the training phase.
	In meta-training, a good initialization $\theta_0$ is pursued so that the target model, starting from $\theta_0$ and applying one or a few iterations of update function $U$, can quickly adapt to a new scenery with limited data samples.
	We adopt the gradient-descent style update function~\cite{li2017meta,park2018meta} which is parameterized by $\alpha$. Then the function $U$ is formulated as:
	\begin{equation}
	U(\theta, \nabla_{\theta}{\mathcal L}; \alpha) = \theta - \alpha \odot \nabla_{\theta}{\mathcal L}.
	\end{equation}
	${\mathcal L}$ is the designed loss function (Eq.~\ref{Overall_L}) for the target model. $\odot$ denotes the element-wise product. $\alpha$ is the parameter that controls the step size of one update iteration, and it is set to the same size as parameter set $\theta$.
	
	To ensure the robustness of scene adaption, during meta-training, the target model is updated and supervised based on the error signals from different input$\&$output pairs in one scene.
	The key idea is that the target model should also generalize to other frames in the same scene, not only several frames which the model is trained on.
	Given a random input$\&$output pair $(x_k, y_k)$ from a normal video, one update step of the target model with initialization $\theta_0$ is derived as: 
	\begin{equation}
	\label{up}
	\theta_0^{i+1} = U(\theta_0^{i}, \nabla_{\theta_0^{i}}{\mathcal L}({y_k},f_{\theta_0}(E_{\eta}({{x}_k})))).
	\end{equation}
	After $T$ update iterations, scene-adapted model parameters $\hat{\theta}$ are obtained. We denote the round of $T$ update iterations as an episode. The iterations number $T$ in an episode is set to 1, to guarantee a fast adaption capability.
	Then we evaluate the model with $\hat{\theta}$ to minimize the scene error signal by running the network through a randomly sampled input$\&$output pair $(x_j,y_j)$ in the same scene as $(x_k,y_k)$.
	
	The gradients of function of gradients algorithm \cite{metz2016unrolled,maclaurin2015gradient,finn2017model,park2018meta} is applied to compute the gradients of above objective function for obtaining a good initialization model $\theta_0^*$ and update step size $\alpha^*$ as:
	\begin{equation}
	\label{su}
	\theta_0^{*},\alpha^* = \argmin_{\theta_0,\alpha} {\mathbb E}[{\mathcal L}({ y_j,f_{\hat{\theta}}(E_{\eta}({{x}_j}))})].
	\end{equation}
	\subsection{Video Anomaly Detection Pipeline}
	\label{vad_pipe}
	We first explain the details of the whole network architecture and how anomaly scores are generated. Then we describe the training and testing phases of our framework.
	
	\noindent \textbf{Network Architecture Details.} 
	Our framework is implemented as a single end-to-end network illustrated in Fig.~\ref{fig:framework}.
	We adopt the same network architecture in~\cite{liu2018future,park2020learning} as the backbone of AE to facilitate a fair comparison.
	In the DPU module, $ M$ attention mapping functions are implemented as fully connected layers to generate a series of normalcy maps and further to form a pool of dynamic prototypes.
	The output encoding of DPU is put forward through the decoder of AE for frame prediction.
	In addition, the DPU module is meta-trained as a few-shot learner, \ie Meta Prototype Unit (MPU). The details are explained below.
	
	\noindent \textbf{Anomaly Score.} To better quantify the anomalous extent of a video frame during inference, we investigate the two cues of feature reconstruction and frame prediction. 
	Since the normal dynamics items in the dynamic prototype pool are learned to encode the compact representations of the normal encoding as in Eq.~\ref{fea_re}, 
	during inference, an anomaly score can be naturally obtained by measuring the compactness error of feature reconstruction term as: ${\mathcal S}_\mathrm{fea} = \mathcal{L}_\mathrm{c}({\mathcal X}_{\rm t}, {\mathcal P}_{\rm t})$. ${\mathcal X}_{\rm t}$ and ${\mathcal P}_{\rm t}$ denote the input encoding map and the dynamic prototype pool of the $\rm t$-th moment, respectively.
	As in previous methods~\cite{liu2018future,gong2019memorizing,park2020learning}, frame prediction error is also leveraged as an anomaly descriptor:
	${\mathcal S}_\mathrm{fra} = \mathcal{L}_\mathrm{fra}(\hat y_{\rm t}, y_{\rm t})$.
	Thus we obtain above two kinds of anomaly scores and combine them with a balance weight $\lambda_{\mathrm s}$ as: ${\mathcal S} = {\mathcal S}_\mathrm{fra} + \lambda_{\mathrm s}{\mathcal S}_\mathrm{fea}$.
	
	\noindent \textbf{Training Phase.} Before meta-training, the AE backbone is first pre-trained using only frame prediction loss (Eq.~\ref{FPL}).
	Then, in a meta-training episode, we randomly sample $K$ tuples of double input$\&$output pairs $\{[(x_i, y_i),(x_j, y_j)]_{i \neq j}\}_{k =1}^{K}$ from a video -- $K$-shot, for parameter update in Eq.~\ref{up} and signal backward in Eq.~\ref{su}.
	Multiple episodes with $K$-shot data sampled from different videos are constructed as a training mini-batch.
	After several times of training epochs with frame pairs sampled from videos of diverse scenes, an initialization parameter set $\theta_0^*$ is obtained, ready for scene adaption.
	
	\noindent \textbf{Testing Phase.} In the testing phase, given a new test sequence, we simply use the first several frames of the sequence to construct $K$-shot input$\&$output frame pairs for updating model parameters. The same procedure is used in the meta-training phase. The updated model is used for detecting anomalies afterwards.
	
	\section{Experiments}
	\subsection{Problem Settings, Datasets and Setups}
	\label{Dataset&Setup}
	\noindent \textbf{Problem Settings.} For better evaluating the effectiveness of our approach, we follow two anomaly detection problem settings, which are the unsupervised setting and few-shot setting. 
	The first one is widely adopted in existing literature~\cite{park2020learning,gong2019memorizing,liu2018future,li2013anomaly,lu2013abnormal,luo2017revisit}, where only normal videos are available during training. The trained models are used to detect anomalies in test videos.
	Note that the scenarios of test videos are seen during training in this setting. 
	The second one, for meta-learning evaluation, is based on collecting training and testing videos from different datasets to make sure the diversity of scenarios during training and testing. This setting is also called `cross-dataset' testing in \cite{lu2020few}.
	In summary, the first setting challenges the approaches for how well they can perform under one fixed camera, while the latter setting examines the adaption capability, when given a new camera.
	We believe above settings are essential for evaluating a robust and practical anomaly detection method.
	
	\noindent \textbf{Datasets.}
	Four popular anomaly detection datasets are selected to evaluate our approach under different problem settings. 
	1) The UCSD Ped1 \& Ped2 dataset \cite{li2013anomaly} contains 34 and 16 training videos, 36 and 12 test videos, respectively, with 12 irregular events, including riding a bike and driving a vehicle. 
	2) The CUHK Avenue dataset \cite{lu2013abnormal} consists of 16 training and 21 test videos with 47 abnormal events such as running and throwing stuff. 
	3) The ShanghaiTech dataset \cite{luo2017revisit} contains 330 training and 107 test videos of 13 scenes.
	4) The UCF-Crime dataset \cite{sultani2018real} contains normal and crime videos collected from a large number of real-world surveillance cameras where each video comes from a different scene.
	We use the 950 normal videos from this dataset for meta-training, then test the model on other datasets in the cross-dataset testing as in ~\cite{lu2020few}.
	
	\noindent \textbf{Evaluation Metrics.}
	Following prior works~\cite{liu2018future,luo2017remembering,mahadevan2010anomaly}, we evaluate the performance using the area under ROC curve (AUC). 
	ROC curve is obtained by varying the threshold for the anomaly score for each frame-wise prediction.
	
	\noindent \textbf{Implementation Details.} Input frames are resized to the resolution of $256 \times 256$ and normalized to the range of $[-1,1]$. During the AE pre-training, the model is trained with the learning rate as $0.0001$ and batch size as $4$.
	In the default setting, DPU is plugged into the AE after the third CNN layer counting backwards, with the encoding feature map of resolution $256 \times 256 \times 128$.
	Training epochs are set to $60, 60,60,10$ on Ped1, Ped2, Avenue and Shanghai Tech, respectively. During meta training, the AE backbone is frozen, only the few-shot target model MPU is trained. The learning rate of the update iteration of the MPU parameter set $\theta$ is set to $0.00001$ for $1000$ training epochs. The mini-batch is set as 10 episodes, and the learning rate of step size $\alpha$ is $0.00001$.
	The balance weights in the objective functions are set as $\lambda_{1} =1$, $\lambda_{2} =0.01$. The desired margin $\gamma$ in feature diversity term is set to $1$. Finally, the hyper-parameter $\lambda_{\mathrm s}$ is set to $1$.
	The experiments are conducted with four Nvidia RTX-2080Ti GPUs. 
	\begin{table}[t]
		\captionsetup{font={small}}
		\small
		\begin{center}
			\caption{Quantitative comparison with state-of-the-art methods for anomaly detection. We measure the average AUC~(\%) on UCSD Ped1 \& Ped2~\cite{li2013anomaly}, CUHK Avenue~\cite{lu2013abnormal}, and ShanghaiTech~\cite{luo2017revisit} in the unsupervised setting. Numbers in bold indicate the best performance and underscored ones are the second best.}
			\scalebox{0.9}{
				\begin{tabular}{c | c | c | c | c}
					\hline
					Methods & Ped1 & Ped2 & Avenue & Shanghai \\
					\hline
					 MPPCA~\cite{kim2009observe} & 59.0 & 69.3 & - & - \\
					 MPPC+SFA~\cite{kim2009observe} & 68.8 & 61.3 & - & - \\
					 MDT~\cite{mahadevan2010anomaly} & 81.8 & 82.9 & - & - \\
					 MT-FRCN~\cite{hinami2017joint} & - & 92.2 & - & - \\
					 Unmasking~\cite{tudor2017unmasking} & 68.4 & 82.2 & 80.6 & - \\
					 SDOR~\cite{pang2020self} & 71.7 & 83.2 & - & -\\
					 ConvAE~\cite{hasan2016learning} & 75.0 & 85.0 & 80.0 & 60.9 \\
					 TSC~\cite{luo2017revisit} & - & 91.0 & 80.6 & 67.9 \\
					 StackRNN~\cite{luo2017revisit} & - & 92.2 & 81.7 & 68.0 \\
					 Frame-Pred~\cite{liu2018future}  & 83.1 & 95.4 & 85.1 & 72.8 \\
					 AMC~\cite{nguyen2019anomaly} & - & 96.2 & 86.9 & -\\
					 rGAN*~\cite{lu2020few} & 83.7 & 95.9 & 85.3 & 73.7 \\
					 rGAN~\cite{lu2020few} & \textbf{86.3} & 96.2 & 85.8 & \textbf{77.9} \\
					 MemAE~\cite{gong2019memorizing} & - & 94.1 & 83.3 & 71.2 \\	
					 LMN~\cite{park2020learning} & - & \textbf{97.0}  & \underline{88.5}&  70.5\\
					\hline
					 Ours w/o DPU. & 83.2 & 95.1 & 84.0 & 66.7 \\
					 Ours w DPU.& \underline{85.1} & \underline{96.9}  &  \textbf{89.5}&  \underline{73.8}\\
					\hline
			\end{tabular}}
			\label{table:Comparison}
		\end{center}	
		\vspace{-0.8cm}
	\end{table}
	
	\subsection{Comparisons with SOTA Methods} 
	\label{VsSOTA}
	
	\begin{table*}[t]
		\caption{Comparison of $K$-shot ($K=0, 1,5,10$) scene-adaptive anomaly detection under the cross-dataset testing setting.
			Note that $K=0$ represents the models are only pre-trained without any adaption.
		}
		\label{table:fs}
		\centering
		\small
		Shanghai Tech
		\begin{tabular*}{\textwidth}{c @{\extracolsep{\fill}}cccccccc}
			\hline
			\textbf{Target} & \textbf{Methods} & \textbf{0-shot (K=0)} & \textbf{1-shot (K=1)} & \textbf{5-shot (K=5)} & \textbf{10-shot (K=10)}\\ 
			\hline\hline
			UCSD Ped 1 & rGAN~\cite{lu2020few} (Finetune) & 73.1 & 76.99 & 77.85  & 78.23 \\ 
			& rGAN~\cite{lu2020few} (Meta) &73.1 & \textbf{80.6} & \textbf{81.42}  & \textbf{82.38} \\
			& \textbf{Ours} (Meta) & \textbf{74.45} & 78.54 & 79.35  & 80.20 \\ 
			\hline
			UCSD Ped 2& rGAN~\cite{lu2020few} (Finetune) & 81.95 & 85.64 & 89.66 & 91.11\\
			& rGAN~\cite{lu2020few} (Meta) & 81.95 & 91.19 & 91.8 & 92.8 \\
			& \textbf{Ours} (Meta) & \textbf{90.17} & \textbf{94.46} & \textbf{94.67}  & \textbf{95.75} \\ 
			\hline
			CUHK Avenue & rGAN~\cite{lu2020few} (Finetune) & 71.43 & 75.43 & 76.52 & 77.77\\
			& rGAN~\cite{lu2020few} (Meta) & 71.43 & 76.58  & 77.1   & 78.79 \\
			& \textbf{Ours} (Meta) & \textbf{74.06} & \textbf{78.92} & \textbf{80.25}  & \textbf{81.69} \\ 
			\hline
		\end{tabular*}
	\end{table*}
	
	\begin{table*}[t!]
		\centering
		\small
		UCF crime
		\begin{tabular*}{\textwidth}{c @{\extracolsep{\fill}}cccccccc}
			\hline
			\textbf{Target} & \textbf{Methods} & \textbf{0-shot (K=0)} & \textbf{1-shot (K=1)} & \textbf{5-shot (K=5)} & \textbf{10-shot (K=10)}\\ 
			\hline\hline
			UCSD Ped 1 & rGAN~\cite{lu2020few} (Finetune) & 66.87 & 71.7 & 74.52  & 74.68 \\ 
			& rGAN~\cite{lu2020few} (Meta) & 66.87 & \textbf{78.44} & \textbf{81.43}  & \textbf{81.62} \\
			& \textbf{Ours} (Meta) & \textbf{75.52} & 77.19 & 78.33  & 79.53 \\ 
			\hline
			UCSD Ped 2& rGAN~\cite{lu2020few} (Finetune) & 62.53 & 65.58 & 72.63 & 78.32 \\
			& rGAN~\cite{lu2020few} (Meta) & 62.53 & 83.08 & 86.41 & \textbf{90.21} \\ 
			& \textbf{Ours} (Meta) & \textbf{86.04} & \textbf{88.43} & \textbf{87.83}  & 89.89 \\ 
			\hline
			CUHK Avenue & rGAN~\cite{lu2020few} (Finetune) & 64.32 & 66.7 & 67.12 & 70.61\\
			& rGAN~\cite{lu2020few} (Meta) & 64.32 & 72.62  & 74.68   & 79.02 \\
			& \textbf{Ours} (Meta) & \textbf{82.26} & \textbf{85.62} & \textbf{85.66}  & \textbf{85.91} \\ 
			\hline
		\end{tabular*}
	\end{table*}
	
	\noindent \textbf{Evaluation under the unsupervised setting.} We first perform an experiment to show that our proposed backbone architecture is comparable to the state-of-the-arts. Note that this sanity check uses the standard training/test setup (training set and testing set are provided by the original datasets), and our model can be directly compared with other existing methods.
	Table~\ref{table:Comparison} shows the comparisons among our proposed architecture and other methods when using the standard unsupervised anomaly detection setup on several anomaly detection datasets. 
	MemAE~\cite{gong2019memorizing} and LMN~\cite{park2020learning} are most-related methods to our approach. 
	They learn a large memory bank for storing normal patterns across the training videos.
	While we propose to learn a few dynamic normal prototypes conditioned on input data, which is more memory-efficient. 
	The superior performance also demonstrates the effectiveness of our DPU module.
	On ped1 and Shanghai Tech, AUCs of our approach are lower than those of rGAN~\cite{lu2020few}. 
	This is reasonable because the model architecture of rGAN is more complicated.
	rGAN uses a ConvLSTM to retain historical information by stacking AE several times. However, we only apply a single AE. 
	
	\noindent \textbf{Evaluation under the few-shot setting.}
	To demonstrate the scene adaption capacity of our approach, we conduct cross-dataset testing by meta-training on the training set of Shanghai Tech and normal videos of UCF-Crime, and then using the other datasets (UCSD Ped1, UCSD Ped2, CUHK Avenue) for validation.
	The comparison results are reported in Table~\ref{table:fs}. 
	As we can see, on most circumstances, the pre-trained DPU model is more generalizing than rGAN.
	Feature reconstruction based on prototypes largely boosts the robustness of anomaly detection with frame prediction.
	Furthermore, $4 \sim 5$\% gain can be achieved with our MPU ($10$-shot to $0$-shot) on various benchmarks. 
	The performance of our MPU-based AE is superior/comparable to the SOTA few-shot learner (rGAN (Meta))~\cite{lu2020few}, with a significantly faster adaption and inference speed.
	We provide more detailed model complexity and inference speed in Sec.~\ref{MC&IS}.
	\begin{figure}
		\centering
		\includegraphics[width=0.45\textwidth]{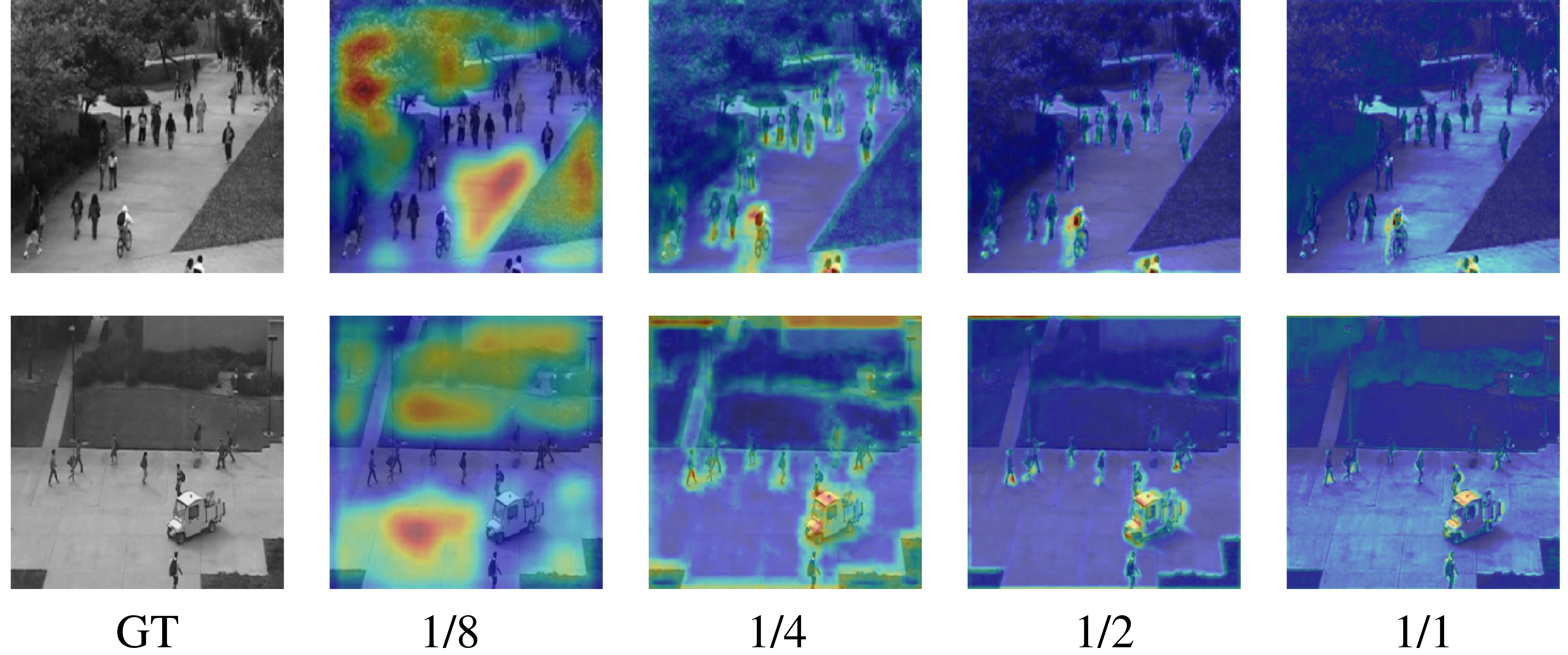}
		\vspace{-5pt}
		\caption{Visualization of AE encoding activation maps from the perspective of L2\_norm. GT stands for ground-truth frame and the annotations of other columns denote the corresponding ratios of input images resolution ($256\times 256$).} 
		\label{l2}
	\end{figure}

	\subsection{Model Complexity and Inference Speed}
	\label{MC&IS}
	With a single Nvidia RTX-2080Ti GPU, our model can run at 166.8 FPS. Note that our DPU module only consumes 1.28K extra parameters (with 10 prototypes).
	Although the parameter size of MemAE~\cite{gong2019memorizing} is smaller than that of ours, the large memory bank used in MemAE leads to a time-consuming read operation, so as the whole inference procedure. Apart from model parameters, our model does not need extra memory space for prototypes, which can be viewed as latent feature vectors.
	Moreover, the inference speed of our method is almost 80 $\times$ faster than rGAN~\cite{lu2020few}. The update iteration for scene adaption of our model ($K=1$) takes only $0.04$ seconds (23.9 FPS). This is almost 19 $\times$ faster than rGAN~\cite{lu2020few} ($K=1$) which takes 0.75 seconds (1.3 FPS). The fast inference speed makes our method more favorable in real-world applications.
	
	\begin{table}[t]
		\captionsetup{font={small}}
		\small
		\begin{center}
			\caption{Analysis on the model complexity and inference speed of various SOTA methods. The inference speed information is collected by running the official implements on a single Nvidia RTX-2080Ti GPU on a machine with 4 CPU cores of E5-2650 v4@2.20GHz and 27.5 G memory.}
			\begin{tabular}{c | c | c}
				\hline
				Methods & Parameters (M) & FPS \\
				\hline
				rGAN~\cite{lu2020few}& 19.0 & 2.1 \\
				MemAE~\cite{gong2019memorizing} & 6.2 & 86.7 \\
				LMN~\cite{park2020learning}& 15.0 &  126.3\\
				\hline
				Ours& 12.7 &  166.8\\
				\hline
			\end{tabular}
			\label{table:Params&Speed}
		\end{center}	
		\vspace{-0.5cm}
	\end{table}
	\begin{figure*}
		\centering
		\includegraphics[width=0.85\textwidth]{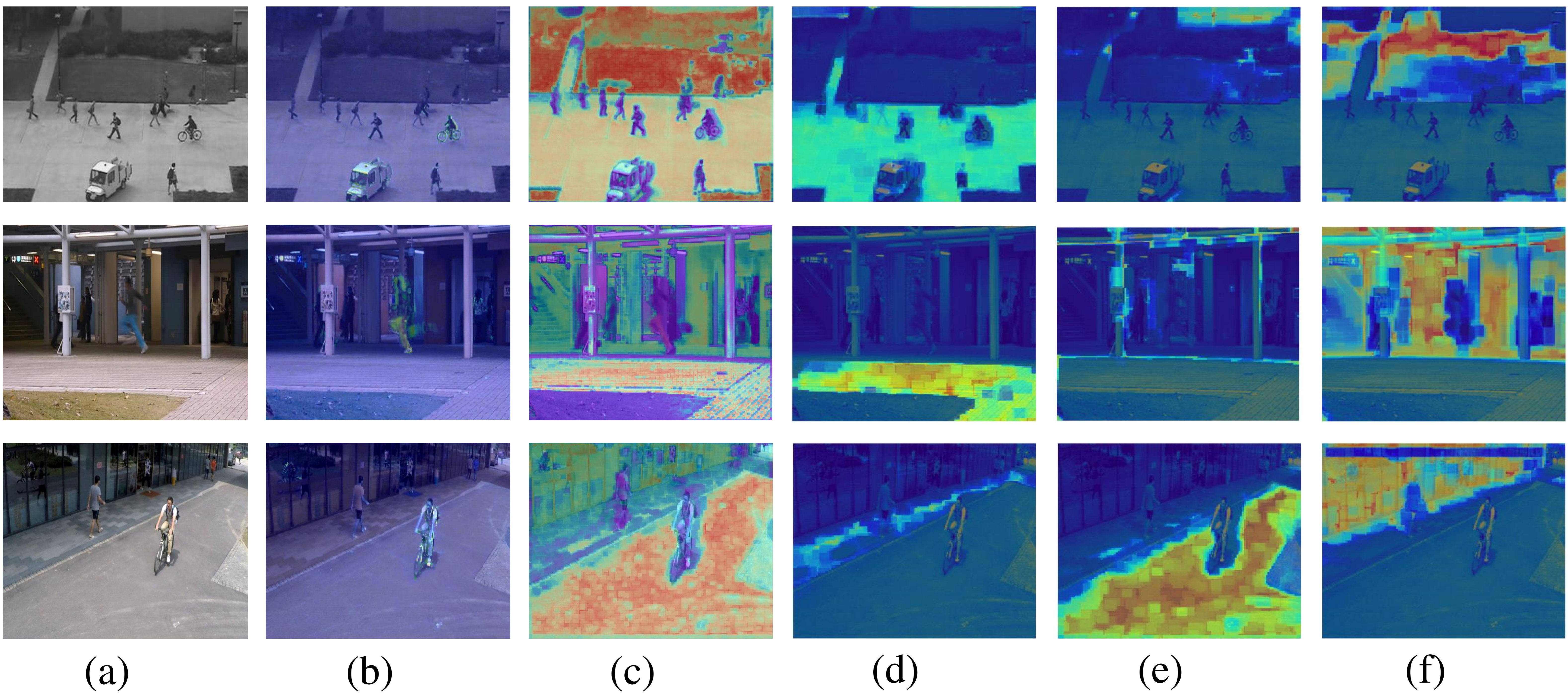}
		\vspace{-7pt}
		\caption{Visualization of some examples of test cases and DPU normalcy maps. The groups of pictures in different columns denote (a) ground-truth frame, (b) error map, (c) sum of normalcy map in DPU, (d) $\sim$ (f) various normalcy map, respectively.} 
		\label{Output}
		\vspace{-0.5cm}
	\end{figure*}

	\subsection{Ablation Studies}
	\label{Ablation}
	\noindent \textbf{Model Component Analysis.} We first analyze the effectiveness of DPU. We set $ M = 10 $ as the default number of the attention mapping functions in DPU.
	The results are listed in Table~\ref{table:MCA}.
	It is clear that the overall performances on various benchmarks are boosted with our DPU by a large margin. 
	We also visualize some example prediction error maps as well as the normalcy maps in DPU in Fig.~\ref{Output}.
	To better analyze the learned normalcy maps, we aggregate all $M$ maps by the sum operation as in Fig.~\ref{Output} (c).
	The normalcy maps encode diverse normal attributes of the scenes such as roads, grasses, and buildings, shown in the columns of (d) $\sim$ (f).
	Furthermore, the weights in suspicious regions are far smaller than those in other parts of the map, indicating that the normal patterns are well encoded as prototypes.
	
	\noindent \textbf{DPU Resolution Analysis.} To investigate the effect of the plugging spot of the DPU module, we carry out experiments on four positions with encoding maps of different resolutions.
	The results are listed in Table~\ref{table:RA}. The AUC results are derived from the feature reconstruction anomaly score on Ped2 dataset. The performance increases along with the resolution.
	We visualize the activation map of the encoding using the L2-norm of the spatial encoding vectors in Fig.~\ref{l2}.
	The higher the activation value, the more information is included in the encoding vector.
	We find that in the higher resolution layers of AE, more anomaly cues are included, which is beneficial for measuring the anomalous extent with the feature reconstruction.
	\label{MPU_R}
	
	\begin{table}[t]
		\captionsetup{font={small}}
		\small
		\begin{center}
			\caption{AUC analysis of the designed DPU module. In the table, FR and FP stand for anomaly scores derived from the Feature Reconstruction and the Frame Prediction, respectively.}
			\scalebox{0.85}{
				\begin{tabular}{c | c  c  c  c}
					\hline
					Setting & Shanghai & Avenue & Ped2 & Ped1\\
					\hline
					AE baseline (FP)& 66.7 & 83.9 & 95.1 & 83.2\\
					AE with DPU (FP) & 71.1 & 85.2 & 92.6 & 83.5\\
					AE with DPU (FR)& 71.9 & 87.1 & 96.2 & 74.1\\
					\hline
					AE with DPU (FP \& FR)& \bf{73.8} & \bf{89.5} & \bf{96.9} & \bf{85.1}\\
					\hline
			\end{tabular}}
			\label{table:MCA}
		\end{center}	
		\vspace{-0.5cm}
	\end{table}
	\begin{table}[t]
		\captionsetup{font={small}}
		\small
		\begin{center}
			\caption{Analysis on the plugging spot of the DPU module. The resolution is divided by the resolution of input images ($256 \times 256$).}
			\begin{tabular}{c | c  c  c  c}
				\hline
				Resolution & \bf{1/1} & 1/2 & 1/4 & 1/8\\
				\hline
				AUC& \bf{89.19} & 86.72 & 84.66 & 81.18\\
				\hline
			\end{tabular}
			\label{table:RA}
		\end{center}	
		\vspace{-0.8cm}
	\end{table}
	\begin{table}[t]
		\captionsetup{font={small}}
		\small
		\begin{center}
			\caption{AUC analysis on the quantity of prototypes in DPU.}
			\begin{tabular}{c | c c c  c c}
				\hline
				Number & 1 & 5 & \bf{10}  & 20 & 40\\
				\hline
				FR & 87.69 & 90.49 & \bf{92.59} & 88.26& 84.37\\
				FP & 94.86 & 95.45 & \bf{96.22} & 95.70& 95.11\\
				\hline
				Overall & 95.22 & 95.57 & \bf{96.90} & 96.03 & 95.74\\
				\hline
			\end{tabular}
			\label{table:PQ}
		\end{center}	
		\vspace{-0.8cm}
	\end{table}
	
	\noindent \textbf{Prototype Quantity Analysis.} To encode the normal dynamics as prototypes, we propose to leverage multiple attention mapping functions for measuring the normalcy of encoding vectors and deriving prototypes as ensembles of the vectors. The number of the attention functions, also denoting the quantity of prototypes, serves as the up-bound of the diverse prototypes needed in one scenario. Experimental results on Ped2 are in Table~\ref{table:PQ}. Based on the results, $M=10$ is an appropriate number of required prototypes. With the number increasing, more noise information is involved and the diversity of prototype items can not be guaranteed, leading to a drastic decline of the performance.
	
	\section{Conclusion}
	In this work, we have introduced a prototype learning module to explicitly model the normal dynamics in video sequences with an attention mechanism for unsupervised anomaly detection.
	The prototype module is fully differentiable and trained in an end-to-end manner. 
	Without extra memory consumption, our approach achieves SOTA performance on various anomaly detection benchmarks in the unsupervised setting.
	In addition, we improve the prototype module as a few-shot normalcy learner with the meta-learning technology. 
	Extensive experimental evaluations demonstrate the efficiency of the scene-adaption approach.
	
	\vspace{5pt}
	\noindent \small{\textbf{Acknowledgments.} This work was supported by the National Natural Science Foundation of China (Grants Nos. 62072244, 61972204, 61906094), the Natural Science Foundation of Jiangsu Province (Grant No. BK20190019).}

	{\small
		\bibliographystyle{ieee_fullname}
		\bibliography{egbib_2021}

\begin{thebibliography}{10}\itemsep=-1pt

\bibitem{abati2019latent}
Davide Abati, Angelo Porrello, Simone Calderara, and Rita Cucchiara.
\newblock Latent space autoregression for novelty detection.
\newblock In {\em CVPR}, 2019.

\bibitem{adam2008robust}
Amit Adam, Ehud Rivlin, Ilan Shimshoni, and Daviv Reinitz.
\newblock Robust real-time unusual event detection using multiple
  fixed-location monitors.
\newblock {\em TPAMI}, 2008.

\bibitem{benezeth2009abnormal}
Yannick Benezeth, P-M Jodoin, Venkatesh Saligrama, and Christophe Rosenberger.
\newblock Abnormal events detection based on spatio-temporal co-occurences.
\newblock In {\em CVPR}, 2009.

\bibitem{chalapathy2017robust}
Raghavendra Chalapathy, Aditya~Krishna Menon, and Sanjay Chawla.
\newblock Robust, deep and inductive anomaly detection.
\newblock In {\em Joint European Conference on Machine Learning and Knowledge
  Discovery in Databases}, 2017.

\bibitem{chandola2009anomaly}
Varun Chandola, Arindam Banerjee, and Vipin Kumar.
\newblock Anomaly detection: A survey.
\newblock {\em ACM computing surveys (CSUR)}, 2009.

\bibitem{chong2017abnormal}
Yong~Shean Chong and Yong~Haur Tay.
\newblock Abnormal event detection in videos using spatiotemporal autoencoder.
\newblock In {\em International Symposium on Neural Networks}, 2017.

\bibitem{cong2011sparse}
Yang Cong, Junsong Yuan, and Ji Liu.
\newblock Sparse reconstruction cost for abnormal event detection.
\newblock In {\em CVPR}, 2011.

\bibitem{finn2017model}
Chelsea Finn, Pieter Abbeel, and Sergey Levine.
\newblock Model-agnostic meta-learning for fast adaptation of deep networks.
\newblock In {\em ICML}, 2017.

\bibitem{fu2019dual}
Jun Fu, Jing Liu, Haijie Tian, Yong Li, Yongjun Bao, Zhiwei Fang, and Hanqing
  Lu.
\newblock Dual attention network for scene segmentation.
\newblock In {\em CVPR}, 2019.

\bibitem{gong2019memorizing}
Dong Gong, Lingqiao Liu, Vuong Le, Budhaditya Saha, Moussa~Reda Mansour, Svetha
  Venkatesh, and Anton van~den Hengel.
\newblock Memorizing normality to detect anomaly: Memory-augmented deep
  autoencoder for unsupervised anomaly detection.
\newblock In {\em ICCV}, 2019.

\bibitem{hasan2016learning}
Mahmudul Hasan, Jonghyun Choi, Jan Neumann, Amit~K Roy-Chowdhury, and Larry~S
  Davis.
\newblock Learning temporal regularity in video sequences.
\newblock In {\em CVPR}, 2016.

\bibitem{hinami2017joint}
Ryota Hinami, Tao Mei, and Shin'ichi Satoh.
\newblock Joint detection and recounting of abnormal events by learning deep
  generic knowledge.
\newblock In {\em CVPR}, 2017.

\bibitem{hu2018squeeze}
Jie Hu, Li Shen, and Gang Sun.
\newblock Squeeze-and-excitation networks.
\newblock In {\em CVPR}, 2018.

\bibitem{kim2009observe}
Jaechul Kim and Kristen Grauman.
\newblock Observe locally, infer globally: a space-time mrf for detecting
  abnormal activities with incremental updates.
\newblock In {\em CVPR}, 2009.

\bibitem{5206569}
J. Kim and K. Grauman.
\newblock Observe locally, infer globally: A space-time mrf for detecting
  abnormal activities with incremental updates.
\newblock In {\em CVPR}, 2009.

\bibitem{kligvasser2018xunit}
Idan Kligvasser, Tamar Rott~Shaham, and Tomer Michaeli.
\newblock xunit: Learning a spatial activation function for efficient image
  restoration.
\newblock In {\em CVPR}, 2018.

\bibitem{koch2015siamese}
Gregory Koch, Richard Zemel, and Ruslan Salakhutdinov.
\newblock Siamese neural networks for one-shot image recognition.
\newblock In {\em ICML deep learning workshop}, 2015.

\bibitem{lake2015human}
Brenden~M Lake, Ruslan Salakhutdinov, and Joshua~B Tenenbaum.
\newblock Human-level concept learning through probabilistic program induction.
\newblock {\em Science}, 2015.

\bibitem{li2013anomaly}
Weixin Li, Vijay Mahadevan, and Nuno Vasconcelos.
\newblock Anomaly detection and localization in crowded scenes.
\newblock {\em TPAMI}, 2013.

\bibitem{li2019selective}
Xiang Li, Wenhai Wang, Xiaolin Hu, and Jian Yang.
\newblock Selective kernel networks.
\newblock In {\em CVPR}, 2019.

\bibitem{li2017meta}
Zhenguo Li, Fengwei Zhou, Fei Chen, and Hang Li.
\newblock Meta-sgd: Learning to learn quickly for few-shot learning.
\newblock {\em ArXiv}, 2017.

\bibitem{liu2018future}
Wen Liu, Weixin Luo, Dongze Lian, and Shenghua Gao.
\newblock Future frame prediction for anomaly detection--a new baseline.
\newblock In {\em CVPR}, 2018.

\bibitem{lu2013abnormal}
Cewu Lu, Jianping Shi, and Jiaya Jia.
\newblock Abnormal event detection at 150 fps in matlab.
\newblock In {\em ICCV}, 2013.

\bibitem{lu2019future}
Yiwei Lu, Mahesh Kumar Krishna~Reddy, Seyed~shahabeddin Nabavi, and Yang Wang.
\newblock Future frame prediction using convolutional vrnn for anomaly
  detection.
\newblock In {\em AVSS}, 2019.

\bibitem{lu2020few}
Yiwei Lu, Frank Yu, Mahesh Kumar~Krishna Reddy, and Yang Wang.
\newblock Few-shot scene-adaptive anomaly detection.
\newblock In {\em ECCV}, 2020.

\bibitem{luo2017remembering}
Weixin Luo, Wen Liu, and Shenghua Gao.
\newblock Remembering history with convolutional lstm for anomaly detection.
\newblock In {\em ICME}, 2017.

\bibitem{luo2017revisit}
Weixin Luo, Wen Liu, and Shenghua Gao.
\newblock A revisit of sparse coding based anomaly detection in stacked rnn
  framework.
\newblock In {\em ICCV}, 2017.

\bibitem{lv2020localizing}
Hui Lv, Chuanwei Zhou, Chunyan Xu, Zhen Cui, and Jian Yang.
\newblock Localizing anomalies from weakly-labeled videos.
\newblock {\em ArXiv}, 2020.

\bibitem{maclaurin2015gradient}
Dougal Maclaurin, David Duvenaud, and Ryan Adams.
\newblock Gradient-based hyperparameter optimization through reversible
  learning.
\newblock In {\em ICML}, 2015.

\bibitem{mahadevan2010anomaly}
Vijay Mahadevan, Weixin Li, Viral Bhalodia, and Nuno Vasconcelos.
\newblock Anomaly detection in crowded scenes.
\newblock In {\em CVPR}, 2010.

\bibitem{masci2011stacked}
Jonathan Masci, Ueli Meier, Dan Cire{\c{s}}an, and J{\"u}rgen Schmidhuber.
\newblock Stacked convolutional auto-encoders for hierarchical feature
  extraction.
\newblock In {\em ICANN}, 2011.

\bibitem{metz2016unrolled}
Luke Metz, Ben Poole, David Pfau, and Jascha Sohl-Dickstein.
\newblock Unrolled generative adversarial networks.
\newblock In {\em ICLR}, 2017.

\bibitem{munkhdalai2017meta}
Tsendsuren Munkhdalai and Hong Yu.
\newblock Meta networks.
\newblock In {\em ICML}, 2017.

\bibitem{nguyen2019anomaly}
Trong-Nguyen Nguyen and Jean Meunier.
\newblock Anomaly detection in video sequence with appearance-motion
  correspondence.
\newblock In {\em ICCV}, 2019.

\bibitem{pang2020self}
Guansong Pang, Cheng Yan, Chunhua Shen, Anton van~den Hengel, and Xiao Bai.
\newblock Self-trained deep ordinal regression for end-to-end video anomaly
  detection.
\newblock In {\em CVPR}, 2020.

\bibitem{park2018meta}
Eunbyung Park and Alexander~C Berg.
\newblock Meta-tracker: Fast and robust online adaptation for visual object
  trackers.
\newblock In {\em ECCV}, 2018.

\bibitem{park2020learning}
Hyunjong Park, Jongyoun Noh, and Bumsub Ham.
\newblock Learning memory-guided normality for anomaly detection.
\newblock In {\em CVPR}, 2020.

\bibitem{ronneberger2015u}
Olaf Ronneberger, Philipp Fischer, and Thomas Brox.
\newblock U-net: Convolutional networks for biomedical image segmentation.
\newblock In {\em MICCAI}, 2015.

\bibitem{sabokrou2016video}
Mohammad Sabokrou, Mahmood Fathy, and Mojtaba Hoseini.
\newblock Video anomaly detection and localization based on the sparsity and
  reconstruction error of auto-encoder.
\newblock {\em Electronics Letters}, 2016.

\bibitem{sabokrou2018adversarially}
Mohammad Sabokrou, Mohammad Khalooei, Mahmood Fathy, and Ehsan Adeli.
\newblock Adversarially learned one-class classifier for novelty detection.
\newblock In {\em CVPR}, 2018.

\bibitem{santoro2016meta}
Adam Santoro, Sergey Bartunov, Matthew Botvinick, Daan Wierstra, and Timothy
  Lillicrap.
\newblock Meta-learning with memory-augmented neural networks.
\newblock In {\em ICML}, 2016.

\bibitem{su2019multi}
Kai Su, Dongdong Yu, Zhenqi Xu, Xin Geng, and Changhu Wang.
\newblock Multi-person pose estimation with enhanced channel-wise and spatial
  information.
\newblock In {\em CVPR}, 2019.

\bibitem{sultani2018real}
Waqas Sultani, Chen Chen, and Mubarak Shah.
\newblock Real-world anomaly detection in surveillance videos.
\newblock In {\em CVPR}, 2018.

\bibitem{sung2018learning}
Flood Sung, Yongxin Yang, Li Zhang, Tao Xiang, Philip~HS Torr, and Timothy~M
  Hospedales.
\newblock Learning to compare: Relation network for few-shot learning.
\newblock In {\em CVPR}, 2018.

\bibitem{tudor2017unmasking}
Radu Tudor~Ionescu, Sorina Smeureanu, Bogdan Alexe, and Marius Popescu.
\newblock Unmasking the abnormal events in video.
\newblock In {\em ICCV}, 2017.

\bibitem{vaswani2017attention}
Ashish Vaswani, Noam Shazeer, Niki Parmar, Jakob Uszkoreit, Llion Jones,
  Aidan~N Gomez, {\L}ukasz Kaiser, and Illia Polosukhin.
\newblock Attention is all you need.
\newblock In {\em NeurIPS}, 2017.

\bibitem{vinyals2016matching}
Oriol Vinyals, Charles Blundell, Timothy Lillicrap, Daan Wierstra, et~al.
\newblock Matching networks for one shot learning.
\newblock In {\em Advances in neural information processing systems}, 2016.

\bibitem{wang2017residual}
Fei Wang, Mengqing Jiang, Chen Qian, Shuo Yang, Cheng Li, Honggang Zhang,
  Xiaogang Wang, and Xiaoou Tang.
\newblock Residual attention network for image classification.
\newblock In {\em CVPR}, 2017.

\bibitem{woo2018cbam}
Sanghyun Woo, Jongchan Park, Joon-Young Lee, and In So~Kweon.
\newblock Cbam: Convolutional block attention module.
\newblock In {\em ECCV}, 2018.

\bibitem{yu2018learning}
Changqian Yu, Jingbo Wang, Chao Peng, Changxin Gao, Gang Yu, and Nong Sang.
\newblock Learning a discriminative feature network for semantic segmentation.
\newblock In {\em CVPR}, 2018.

\bibitem{zhao2011online}
Bin Zhao, Li Fei-Fei, and Eric~P Xing.
\newblock Online detection of unusual events in videos via dynamic sparse
  coding.
\newblock In {\em CVPR}, 2011.

\bibitem{zhao2018psanet}
Hengshuang Zhao, Yi Zhang, Shu Liu, Jianping Shi, Chen Change~Loy, Dahua Lin,
  and Jiaya Jia.
\newblock Psanet: Point-wise spatial attention network for scene parsing.
\newblock In {\em ECCV}, 2018.

\bibitem{zhong2019graph}
Jia-Xing Zhong, Nannan Li, Weijie Kong, Shan Liu, Thomas~H Li, and Ge Li.
\newblock Graph convolutional label noise cleaner: Train a plug-and-play action
  classifier for anomaly detection.
\newblock In {\em CVPR}, 2019.

\end{thebibliography}
	}
	
\end{document}